\documentclass[11pt]{article}
\usepackage{eacl2017}
\usepackage{times}
\usepackage{url}
\usepackage{latexsym}
\usepackage{amssymb}
\usepackage{amsmath}

\usepackage{todonotes}
\usepackage{graphicx}
\usepackage{subfig}
\usepackage{latexsym}
\usepackage{textcomp}
\usepackage[titletoc,title]{appendix}
\usepackage{xcolor,colortbl}
\usepackage[acronym]{glossaries}
\usepackage{float}
\restylefloat{table}

\newcommand*\rot{\rotatebox{90}}

\def\None{\ensuremath\text{\it None}}

\newacronym{LSTM}{LSTM}{Long Short-Term Memory}
\newacronym{SLU}{SLU}{Spoken Language Understanding}
\newacronym{ASR}{ASR}{Automatic Speech Recognition}
\newacronym{DST}{DST}{Dialog State Tracker}
\newacronym{DSTC}{DSTC}{Dialog State Tracking Challenge}
\newacronym{HIS}{HIS}{Hidden Information State}
\newacronym{CRF}{CRF}{Conditional Random Field}
\newacronym{ME}{ME}{Maximum Entropy}
\newacronym{SDS}{SDS}{Spoken Dialogue System}
\newacronym{NN}{NN}{Neural Network}
\newacronym{HDST}{HDST}{Hybrid Dialog State Tracker}

\newcommand{\ignore}[1]{}

\newcommand{\M}[1]{{\color{black}#1}}
\newcommand{\R}[1]{{\color{black}#1}}

\eaclfinalcopy 
\def\eaclpaperid{381} 

\newcommand{\superscript}[1]{\ensuremath{^{\textrm{#1}}}}
\def\dg{\superscript{\dag}}
\def\ddg{\superscript{\ddag}}

\title{Hybrid Dialog State Tracker with ASR Features}

\author{
Miroslav Vodol{\'a}n\dg\ddg\\
Charles University in Prague\ddg\\
Faculty of Mathematics and Physics\\
Malostranske nam. 25, 11800, Prague\\
  \hskip6.0cm\begin{tabular}{ccc}
    \texttt{\{mvodolan, rudolf\_kadlec, jankle\}@cz.ibm.com} \\
    \texttt{vodolan@ufal.mff.cuni.cz}
  \end{tabular}
\And
Rudolf Kadlec\dg and Jan Kleindienst\dg\\
IBM Watson\dg\\
V Parku 4\\
Prague 4, Czech Republic\\
}

\date{}

\begin{document}
\maketitle
\begin{abstract}
This paper presents a hybrid dialog state tracker enhanced by trainable Spoken Language Understanding (SLU) for slot-filling dialog systems. 
\R{Our architecture is inspired by previously proposed neural-network-based belief-tracking systems.
In addition we extended some parts of our modular architecture with differentiable rules to allow end-to-end training. We hypothesize that these rules allow our tracker to generalize better than pure machine-learning based systems.
}
For evaluation we used the Dialog State Tracking Challenge (DSTC) 2 dataset - a popular belief tracking testbed with dialogs from restaurant information system.
To our knowledge, our hybrid tracker sets a new state-of-the-art result in three out of four categories within the DSTC2.

\end{abstract}

\section{Introduction}
A belief-state tracker is an important component of dialog systems whose responsibility is to predict user's goals based on history of the dialog. 
Belief-state tracking was extensively studied in
the \gls{DSTC} series~\cite{williams2016dialog} by providing shared testbed for various tracking approaches.
The \gls{DSTC} abstracts away the subsystems of end-to-end spoken dialog systems, focusing only on the dialog state tracking. It does so by providing datasets of \gls{ASR} and \gls{SLU} outputs with reference transcriptions, together with annotation on the level of dialog acts and user goals on slot-filling tasks where dialog system tries to fill predefined slots with values from a known ontology (e.g. \textit{moderate} value for a \textit{pricerange} slot).

In this work we improve state-of-the-art results on \gls{DSTC}2~\cite{henderson-thomson-williams:2014:W14-43} by combining two central ideas previously proposed in different successful models: 1) machine learning core with hand-coded\footnote{For historical reasons we adopted the \textit{hand-coded rules} term used throughout the belief tracking community. From another viewpoint, our rules can be seen as a linear combination model.} rules, an idea already explored by~\newcite{Yu2015} and \newcite{hybrid_tracker} with 2) a complex neural network based architecture that processes \gls{ASR} features proposed by ~\newcite{henderson-thomson-young:2014:W14-43}.
Their network consist of two main units. One unit handles generic behaviour that is independent of the actual slot value and the other depends on slot value and can account for common confusions.

When compared to~\newcite{henderson-thomson-young:2014:W14-43} that inspired our work:
1) our model does not require auto-encoder pre-training and shared initial training on all slots
which makes the training easier; 
2) our approach combines a rule-based core of the tracker and RNNs while their model used only RNNs; 
3) we use different NN architecture to process \gls{SLU} features.

In the next section we describe the structure of our model, after that we detail how we evaluated the model on the DSTC2 dataset. 
We close the paper with a section on the lessons we learned.

\section{Hybrid dialog state tracker model}

The tracker operates separately on the probability distribution for each slot. Each turn, the tracker generates these distributions to reflect the user's goals based on the last action of the machine, the observed user actions, the probability distributions from the previous turn and an internal hidden state. The probability distribution $h^s_t[v]$ is a distribution over all possible values $v$ from the domain of slot $s$ at dialog turn $t$.
The joint belief state is represented by a probability distribution over the Cartesian product of the individual slot domains.

\begin{figure}[!htbp]
\begin{center}
    \includegraphics[scale=0.5]{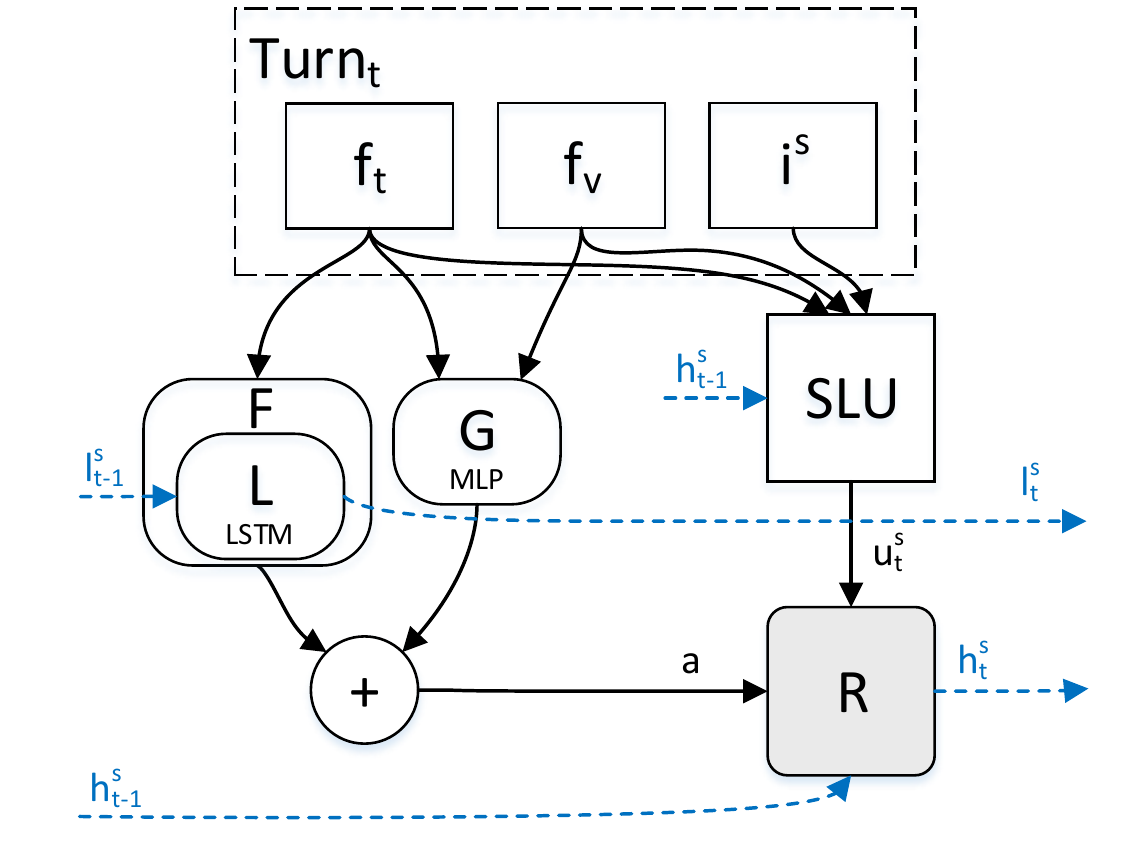} 
\end{center}
    \caption{The structure of the Hybrid tracker at turn $t$. It is a recurrent model that uses the probability distribution $h^s_{t-1}$ and hidden state $l^s_{t-1}$  from the previous turn (recurrent information flow is depicted by dashed blue lines). Inputs of the machine-learned part of the model (represented by functions $G$ and $F$ based on recurrent $L$) are the turn and value features $f_t$, $f_v$ and the hidden state. The features are used to produce transition coefficients $a$ for the $R$ function which transforms the output of the SLU $u^s_t$ into belief $h^s_t$.}
    
    \label{fig:tracker_structure}
\end{figure}

In the following notation $i^s_t$ denotes a user action pre-processed into a probability distribution of informed values for the slot $s$ and turn $t$. During the pre-processing, every \textit{Affirm()} from the \gls{SLU} is transformed into \textit{Inform(s=v)} depending on a machine action of the turn.
The $f_t$ denotes turn features consisting of unigrams, bigrams, and trigrams extracted from the \gls{ASR} hypotheses $N$-best list. They are weighted by the probability of the corresponding hypothesis on the $N$-best list. The same approach is used in~\newcite{henderson-thomson-young:2014:W14-43}. To make our system comparable to the best-performing tracker~\cite{williams:2014:W14-43} we also included features from batch \gls{ASR} (recognition hypotheses and the unigram word-confusion matrix). The batch \gls{ASR} hypotheses are encoded in the same way as hypotheses from the regular \gls{ASR}. The confusion matrix information is encoded as weighted unigrams. The last part of the turn features encodes machine-action dialog acts. We are using trigram-like encoding \textit{dialogact}-\textit{slot}-\textit{value} with weight $1.0$. The other features are value features $f_{v_i}$ created from turn features, which contain  occurrence of $v_i$, by replacing occurrence of the value $v_i$ and slot name $s$ by a common tag \mbox{(\textit{inform-food-italian $\rightarrow$ inform-\textless slot\textgreater-\textless value\textgreater)}}. This technique is called delexicalization by ~\newcite{henderson-thomson-young:2014:W14-43}.

From a high-level perspective, our model consists of a rule-based core represented by a function $R$ that specifies how the belief state evolves based on new observations. The rules $R$ depend on the output of machine-learned SLU and on \emph{transition coefficients}\footnote{These coefficients were modelled by a so called \textit{durability} function in~\newcite{kadlec2014knowledge}.} $a_{v_i,v_j}$ that specify how easy it would be to override a previously internalized slot value $v_j$ with a new value $v_i$ in the given situation. The $a_{v_i,v_j}$ \emph{transition coefficients} are computed as a sum of functions $F$ and $G$ where $F$ accounts for generic value-independent behavior which can however be corrected by the value-dependent function $G$. The structure of the tracker is shown in Figure~\ref{fig:tracker_structure}.

In the next subsection, 
we will describe the rule-based component of the Hybrid tracker. Afterwards, in Section~\ref{ssec:machine_learned_part}, we will describe the machine-learned part of the tracker followed by the description of the trainable SLU in Section~\ref{ssec:slu_part}.

\subsection{Rule-based part}
\label{ssec:knowledge_based_part}

The rule-based part of our tracker, inspired by~\newcite{hybrid_tracker}, is specified by a function $R(h^s_{t-1}, u_t^s, a)=h^s_t$, which is a function of a slot--value probability distribution $h^s_{t-1}$ in the previous turn, the output $u_t^s$ of a trainable \gls{SLU} and of \emph{transition coefficients} $a$ which control how the new belief $h^s_t$ is computed.  The first equation specifies the belief update rule for the probability assigned to slot value $v_i$: 

\begin{equation}
   \label{eq:transition_rules}
    h^s_t[v_i] = h^s_{t-1}[v_i] - \tilde{h}^s_t[v_i] + u^s_t[v_i] \cdot \sum_{v_j \neq v_i}{ h^s_{t-1}[v_j] \cdot a_{v_iv_j} } 
\end{equation}
where $\tilde{h}^s_t[v_i]$ expresses how much probability will be transferred from $h^s_{t-1}[v_j]$ to other slot values in $h^s_{t}$. This is computed as:
\begin{equation}
   \label{eq:transferred_probability}
   \tilde{h}^s_t[v_i] = h^s_{t-1}[v_i]\cdot \sum_{v_j \neq v_i}{u^s_t[v_j] \cdot a_{v_jv_i}}
\end{equation}
where $a_{v_iv_j}$ is called the \emph{transition coefficient} between values $v_i$ and $v_j$. 
These coefficients are computed by the machine-learned part of our model.

\subsection{Machine-learned part}
\label{ssec:machine_learned_part}

The machine-learned part modulates behavior of the rule-based part $R$ by transition coefficients $a_{v_iv_j}$ that control the amount of probability which is transferred from $h^s_{t-1}[v_j]$ to $h^s_t[v_i]$ as in~\newcite{hybrid_tracker}. However, our computation of the coefficients involves two different functions:
\begin{equation}
    a_{v_iv_j} = F(l_{t-1},f_t,v_i,v_j) + G(f_t, v_i,{v_j})
\end{equation}
where the function $F$ controls generic behavior of the tracker, which does not take into account any features about $v_i$ or $v_j$. On the other hand, function $G$ provides value-dependent corrections to the generic behavior described by $F$. 

\textbf{Value Independent Model.}
$F$ is specified as:

\begin{equation}
    F(l_{t-1},f_t,v_i,v_j) = 
     \begin{cases}
        c_{\text{new}} &\text{if~} v_i = \None \\ 
        c_{\text{override}} &\text{if~} v_i \neq v_j \\
   \end{cases}
\end{equation}

where the $F$ function takes values of $c_{\text{new}}$ and $c_{\text{override}}$ from a function $L$. The function $\langle c_{\text{new}}$, $c_{\text{override}}$, $l_t \rangle = L(l_{t-1}, f_t)$ is a recurrent function that takes its hidden state vector $l_{t-1}$ from the previous turn and the turn features $f_t$ as input and it outputs two scalars $c_{\text{new}}$, $c_{\text{override}}$ and a new hidden state $l_t$.
An interpretation of these scalar values is the following:
\begin{itemize}    
    \item $c_{\text{new}}$ --- describes how easy it would be to change the belief from hypothesis \textit{None} to an instantiated slot value,
    \item $c_{\text{override}}$ --- models a goal change, that is, how easily it would be to override the current belief with a new observation.
\end{itemize}

In our implementation, $L$ is formed by 5 LSTM~\cite{Hochreiter1997} cells with $tanh$ activation. We use a recurrent network for $L$ since it can learn to output different values of the $c$ parameters for different parts of the dialog (e.g., it is more likely that a new hypothesis will arise at the beginning of a dialog). This way, the recurrent network influences the rule-based component of the tracker. 
The function $L$ uses the turn features $f_t$, which encode information from the \gls{ASR}, machine actions and the currently tracked slot.

\textbf{Value Dependent Model.} The function $G(f_t, v_i,{v_j})$ corrects the generic behavior of $F$. $G$ is implemented as a multi-layer perceptron \M{with linear activations}, 
that is: $G(f_t, v_i,{v_j}) = MLP(f_t, f_{v_i})|_{v_j}$. The MLP uses turn features $f_t$ together with delexicalized features $f_{v_i}$ for slot value $v_i$. In our implementation the MLP computes a whole vector with values for each $v_k$ at once. However, in this notation we use just the value corresponding to $v_j$. To stress this we use the restriction operator $|_{v_j}$.

\subsection{Spoken Language Understanding part}
\label{ssec:slu_part}
\begin{figure}[!ht]
    \includegraphics[scale=0.5]{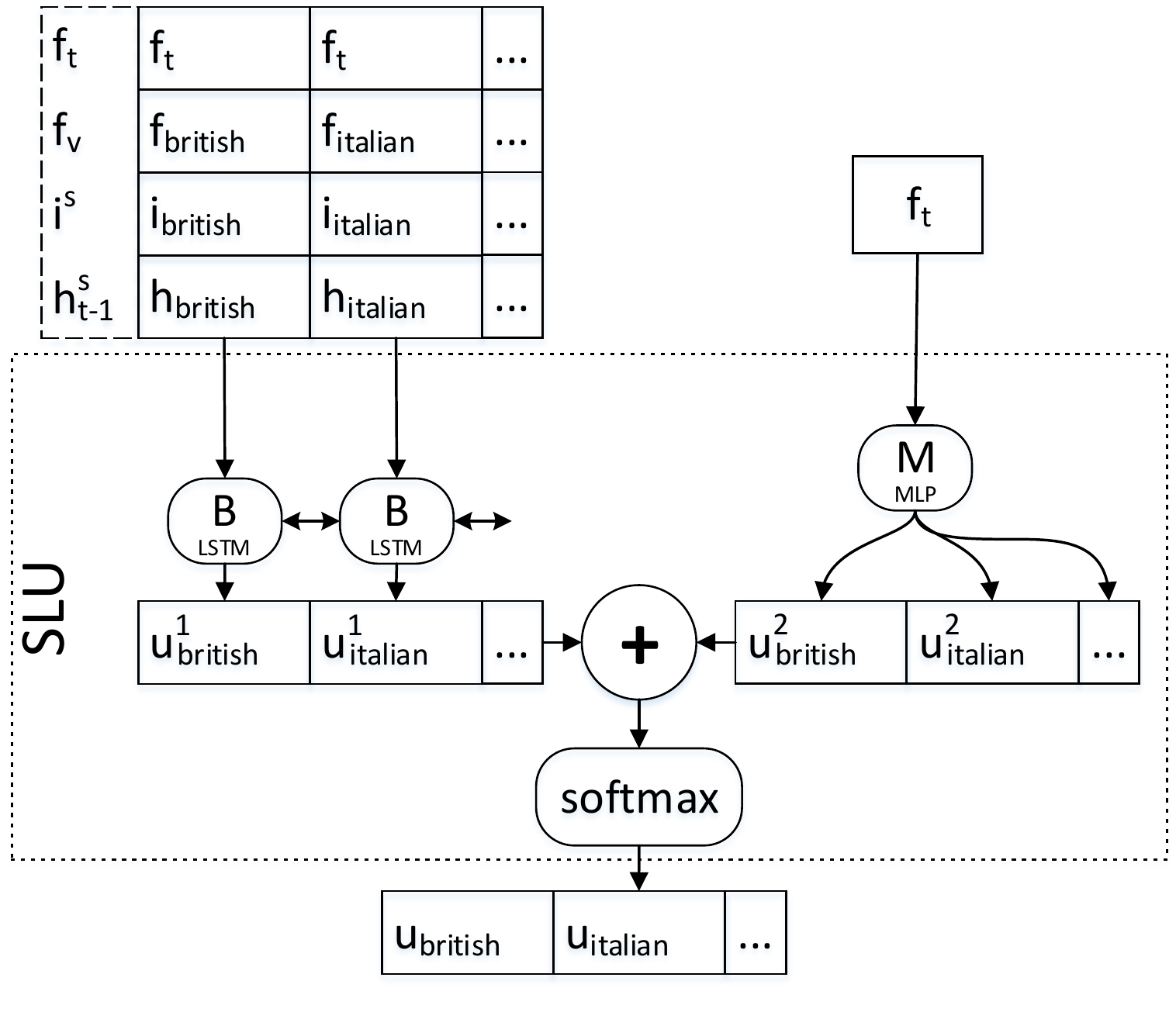}
    \caption{The SLU consists of two units. The first unit processes turn features $f_t$, per-value features $f_v$, original informs $i^s$ and belief from the previous turn $h^s_{t-1}$ by a bidirectional LSTM $B$ and outputs a vector $u^1$. The second unit maps turn features $f_t$ by an MLP $M$ \M{(with two linear hidden layers of sizes 50 and 20 - effect of the first layer is to regularize information passed through the $M$)} onto $u^2$. Softmaxed sum of those output vectors is used as a probability distribution of informed values $u^s_t$.}
    \label{fig:slu_structure} 
\end{figure}
The SLU part of the tracker shown in Figure~\ref{fig:slu_structure} is inspired by an architecture, proposed in~\newcite{henderson-thomson-young:2014:W14-43}, consisting of two separate units. The first unit works with value-independent features $f_{v_i}$ where slot values (like \textit{indian}, \textit{italian}, \textit{north}, etc.) from the ontology are replaced by tags. This allows the unit to work with values that have not been seen during training.

The features are processed by a bidirectional LSTM $B$ (with 10 $tanh$ activated cells) which enables the model to compare the likelihoods of the values in the user utterance. Even though this is not a standard usage of the LSTM it has proved as crucial especially for estimating the \textit{None} value which means that no value from the ontology was mentioned\footnote{We also tested other models, such as max-pooling over feature embeddings (to get extra information for \textit{None} value), however, these performed much worse on the validation dataset.}. The other benefit of this architecture is that it can weight its output $u^1$ according to how many ontology values have been detected during turn $t$.

However, not all ontology values can be replaced by tags because of speech-recognition errors or simply because the ontology representation is not the same as the representation in natural language (e.g. \textit{dontcare}\texttildelow it does not matter). For this purpose, the model uses a second unit that maps untagged features directly into a value vector $u^2$. Because of its architecture, the unit is able to work only with ontology values seen during training.
At the end, outputs $u^1$, $u^2$ of the two units are summed together and turned into a probability distribution $u$ via softmax.
Since all parts of our model ($R$, $F$, $G$, SLU) are differentiable, all parameters of the model can be trained jointly by gradient-descent methods.

\section{Evaluation}

\textbf{Method.} From each dialog in the \textit{dstc2\_train} data ($1612$ dialogs) we extracted training samples for the slots \textit{food}, \textit{pricerange} and \textit{area} and used all of them to train each tracker. The development data \textit{dstc2\_dev} ($506$ dialogs) were used to select the $f_t$ and $f_v$ features. We took the $2000$ most frequent $f_t$ features and the $100$ most frequent $f_v$ features.

The cost that we optimized consists of a tracking cost, which is computed as a cross-entropy between a belief state $h^s_t$ and a goal annotation, and of an \gls{SLU} cost, which is a cross-entropy between the output of the \gls{SLU} $u_t^s$ and a semantic annotation. We did not use any regularization on model parameters. 
We trained the model for 30 epochs by SGD with the AdaDelta~\cite{zeiler2012adadelta} weight-update rule and batch size 16 on fully unrolled dialogs. 
We use the model from the best iteration according to error rate on \textit{dstc2\_dev}.
The evaluated model was an ensemble of 10 best trackers (according to the tracking accuracy on \textit{dstc2\_dev}) selected from $62$ trained trackers. All trackers used the same training settings with difference in initial parameter weights only). 
Our tracker did not track the \textit{name} slot because there are no training data available for it. Therefore, we always set value for the \textit{name} to \textit{None}.

\M{\textbf{Results.} This section briefly summarizes results of our tracker on \textit{dstc2\_test} (1117 dialogs) in all DSTC2 categories as can be seen in Table~\ref{tab:results-dstc2}. We also provide evaluation of the tracker without specific components to measure their contribution in the overall accuracy.

\begin{table*}[!htb]
\small
\begin{center}
\bgroup
\def\arraystretch{1.00}
\setlength{\tabcolsep}{3pt}
\begin{tabular}{ll|cc|cc|c|c}
 & & \multicolumn{5}{c}{dstc2\_test}  \\ \cline{3-8} 
 & & \rot{ASR} & \rot{Batch ASR} & \rot{Accuracy} & \rot{L2} & \rot{post DSTC} & \rot{test validated }\\[0.1cm] \hline

& \textbf{Hybrid Tracker -- this work}  & $\surd$  &  $\surd$ &   \textbf{.810}	& .318 & $\surd$ & $\surd$ \\
& DST2 stacking ensemble \cite{henderson-thomson-williams:2014:W14-43} & $\surd$  & $\surd$ & .798 & \textbf{.308} & $\surd$ & $\surd$ \\ \hline
& \textbf{Hybrid Tracker -- this work}  & $\surd$  &  $\surd$ &   \textbf{.796}	&   \textbf{.338} & $\surd$	\\
& \newcite{williams:2014:W14-43} & $\surd$  &  $\surd$ 	&   .784	&   .735 & &	\\
\hline 
& \textbf{Hybrid Tracker -- this work}  & $\surd$ &  &   \textbf{.780}	&   .356 & $\surd$ \\ 
& \newcite{williams:2014:W14-43} & $\surd$  &  	&   .775	&   .758 & &	\\
& \textbf{Hybrid Tracker without G -- this work}  & $\surd$ &  &  .772	& .368 & $\surd$ \\ 
& \textbf{Hybrid Tracker without M -- this work}  & $\surd$ &  & .770 & .373 & $\surd$ \\ 
& \newcite{henderson-thomson-young:2014:W14-43} \ignore{t4e3} & $\surd$ &  & .768	& \textbf{.346} & & \\
& \textbf{Hybrid Tracker without bidir -- this work}  & $\surd$ &  & .763	&  .375 & $\surd$ \\ 
& \newcite{Yu2015}  & $\surd$ & &  .762	& .436 & $\surd$ & \\
& YARBUS \cite{Fix2015} & $\surd$ & &  .759 & .358 & $\surd$ & \\
& \newcite{sun-EtAl:2014:W14-43} \ignore{t7e4} & $\surd$ &  &   .750 &	.416 & & \\
& Neural Belief Tracker \cite{mrkvsic2016neural} \ignore{t7e4} & $\surd$ &  &   .73? &	??? &  $\surd$ & \\
\hline
& TL-DST \cite{lee2016task} & & & \textbf{.747} & .451 & $\surd$   \\
& \textbf{Hybrid Tracker -- this work}  &  &  &   .746	&  .414 & $\surd$ \\ 

& \newcite{hybrid_tracker} & & & .745 & .433 & $\surd$   \\
& \newcite{williams:2014:W14-43}  & & & .739   &   .721 & 	\\
& \newcite{henderson-thomson-young:2014:W14-43} \ignore{t4e3} & & 	& .737 &	\textbf{.406} &  \\
& Knowledge-based tracker~\cite{kadlec2014knowledge} & & &  .737 & .429 & $\surd$ \\
& \newcite{sun-EtAl:2014:W14-43} \ignore{t7e4}  & &   &   .735   &   .433  & 	 \\
& \newcite{smith:2014:W14-43} \ignore{t3e0}  & & & .729   & .452  & 	 \\
& \newcite{lee-EtAl:2014:W14-43}  & &  & .726 & .427  & 	\\
& YARBUS \cite{Fix2015}  & &  & .725 & .440 & $\surd$  & 	\\
& \newcite{ren-xu-yan:2014:W14-43} \ignore{t6e2}   & &    & .718  &   .437  & 	 & 	 \\
\hline
  & Focus baseline    &  &    &   \textbf{.719}   &   \textbf{.464}   &   \\ 
  & HWU baseline     &  & &   .711  &   .466  & \\ 
\hline

\end{tabular}
\egroup
\end{center}
\caption{
Joint slot tracking accuracy and L2 (denotes the squared L2 norm between the estimated belief distribution and correct distribution) for various systems reported in the literature. The trackers that used ASR/Batch ASR have $\surd$ in the corresponding column. The results of systems that did not participate in DSTC2 are marked by $\surd$ in the ``post DSTC" column. \M{The first group shows results of trackers that used dstc test data for validation}. The second group lists individual trackers that use ASR and Batch ASR features. The third group lists systems that use only the ASR features. The last group lists baseline systems provided by DSTC organizers.}
\label{tab:results-dstc2}
\end{table*}

In the standard categories using Batch ASR and ASR features, we set new state-of-the-art results. In the category without ASR features (SLU only) our tracker is slightly behind the best tracker~\cite{lee2016task}.

For completeness, we also evaluated our tracker in the ``non-standard" category that involves trackers using test data for validation. This setup was proposed in~\newcite{henderson-thomson-williams:2014:W14-43} where an ensemble was trained from all DSTC2 submissions. However, this methodology discards a direct comparison with the other categories since it can overfit to test data.
Our tracker in this category is a weighted\footnote{\M{Validation was used for finding the weights only.}} averaging ensemble of trackers trained for the categories with \gls{ASR} and batch \gls{ASR}.
}

We also tested contribution of specialization components $G$ and $M$ by training new ensembles of models without those components. Accuracy of the ensembles can be seen in Table~\ref{tab:results-dstc2}. From the results can be seen that removing either of the components hurts the performance in a similar way. 

In the last part of evaluation we studied importance of the bidirectional LSTM layer $B$ by ensembling models with linear layer instead. From the table we can see a significant drop in accuracy, showing the $B$ is a crucial part of our model.

\M{
\section{Lessons learned}
\label{sec:model_evolution}

Originally we designed the special \gls{SLU} unit $M$ with a sigmoid activation inspired by architecture of~\cite{henderson-thomson-young:2014:W14-43}. However, we found it difficult to train because gradients were propagated poorly through that layer causing its output to resemble priors of ontology values rather than probabilities of informing some ontology value based on corresponding \gls{ASR} hypotheses as suggested by the network hierarchy. 
The problem resulted in an inability to learn alternative wordings of ontology values which are often present in the training data. One such example can be \textit{``asian food"} which appears 16 times in the training data as a part of the best \gls{ASR} hypothesis while 13 times it really informs about \textit{``asian oriental"} ontology value. Measurements on \textit{dstc2\_dev} have shown that the \gls{SLU} was not able to recognize this alias anytime. 
We managed to solve this training issue by simplifying the special \gls{SLU} sigmoid to linear activation instead.
The resulting \gls{SLU} is able to recognize common alternative wordings as \textit{``asian food"} appearing more than 10 times in training data, as well as rare alternatives like \textit{``anywhere"} (meaning \textit{area:dontcare}) appearing only 5 times in training data.
}

\section{Conclusion}
\label{sec:conslusion}
We have presented an end-to-end trainable belief tracker with modular architecture enhanced by differentiable rules. \M{The modular architecture of our tracker outperforms other approaches in almost all standard DSTC categories without large modifications making our tracker successful in a wide range of input-feature settings.}

\section*{Acknowledgments}
This  work  was  supported by GAUK grant 1170516 of Charles University in Prague.

\bibliography{hybrid_tracker}
\bibliographystyle{eacl2017}

\end{document}